\documentclass[letterpaper]{article} 
\usepackage[]{aaai2026}  
\usepackage{times}  
\usepackage{helvet}  
\usepackage{courier}  
\usepackage[hyphens]{url}  
\usepackage{graphicx} 
\usepackage{natbib}  
\usepackage{caption} 
\usepackage[utf8]{inputenc}

\usepackage{amsmath}
\usepackage{amssymb}
\usepackage{amsthm}
\usepackage{amsfonts}
\usepackage{mathtools}
\usepackage{bm}
\usepackage{nicefrac}

\usepackage{adjustbox}
\usepackage{float}
\usepackage{subfigure}
\usepackage{subfig}
\usepackage{subcaption}
\usepackage{booktabs}
\usepackage{array}
\usepackage{arydshln}
\usepackage{colortbl}
\usepackage{makecell}
\usepackage{multirow}
\usepackage{diagbox}
\usepackage{rotating}

\usepackage{xcolor}
\usepackage{microtype}
\usepackage{tcolorbox}

\usepackage[ruled]{algorithm2e}
\usepackage{algpseudocode}

\usepackage[capitalize,noabbrev]{cleveref}

\usepackage{xpatch}
\usepackage[multiple]{footmisc}

\SetKwComment{Comment}{/* }{ */}
\newtheorem*{theorem*}{Theorem}
\newtheorem*{lemma*}{Lemma}
\theoremstyle{plain}
\newtheorem{theorem}{Theorem}[]

\theoremstyle{definition}

\theoremstyle{remark}

\newcommand{\bd}[1]{\textbf{#1}}

\newcommand{\E}{\mathbb{E}}

\title{SALR: Sparsity-Aware Low-Rank Representation for Efficient Fine-Tuning of Large Language Models}

\author{
    Longteng Zhang\textsuperscript{\rm 1}, Sen Wu\textsuperscript{\rm 3}, Shuai Hou\textsuperscript{\rm 3}, Zhengyu Qing\textsuperscript{\rm 3}, Zhuo Zheng\textsuperscript{\rm 4}, Danning Ke\textsuperscript{\rm 4}, Qihong Lin\textsuperscript{\rm 4},\\
    Qiang Wang\textsuperscript{\rm 3}\footnotemark[1], Shaohuai Shi\textsuperscript{\rm 3}, Xiaowen Chu\textsuperscript{\rm 1,2}\thanks{Corresponding author}\\
}
\affiliations{
    \textsuperscript{\rm 1}The Hong Kong University of Science and Technology (GuangZhou)\\
    \textsuperscript{\rm 2}The Hong Kong University of Science and Technology\\
    \textsuperscript{\rm 3}Harbin Institute of Technology, Shenzhen\\
    \textsuperscript{\rm 4}Huawei Technologies\\
    lzhang330@connect.hkust-gz.edu.cn, \{24S151068, houshuai, 210110609\}@stu.hit.edu.cn\\
    \{zhengzhuo2, kedanning, linqihong\}@huawei.com, 
    \{qiang.wang, shaohuais\}@hit.edu.cn, xwchu@ust.hk
}

\usepackage{bibentry}

\begin{document}

\maketitle

\begin{abstract}
Adapting large pre-trained language models to downstream tasks often entails fine-tuning millions of parameters or deploying costly dense weight updates, which hinders their use in resource-constrained environments. Low-rank Adaptation (LoRA) reduces trainable parameters by factorizing weight updates, yet the underlying dense weights still impose high storage and computation costs. Magnitude-based pruning can yield sparse models but typically degrades LoRA's performance when applied naively. In this paper, we introduce SALR (Sparsity-Aware Low-Rank Representation), a novel fine-tuning paradigm that unifies low-rank adaptation with sparse pruning under a rigorous mean-squared-error framework. We prove that statically pruning only the frozen base weights minimizes the pruning error bound, and we recover the discarded residual information via a truncated-SVD low-rank adapter, which provably reduces per-entry MSE by a factor of $(1 - r/\min(d,k))$. To maximize hardware efficiency, we fuse multiple low-rank adapters into a single concatenated GEMM, and we adopt a bitmap-based encoding with a two-stage pipelined decoding+GEMM design to achieve true model compression and speedup. Empirically, SALR attains 50\% sparsity on various LLMs while matching the performance of LoRA on GSM8K and MMLU, reduces model size by $2\times$, and delivers up to a $1.7\times$ inference speedup.
\end{abstract}

\section{Introduction}

The rapid growth of large-scale neural networks has driven remarkable advances across natural language processing, computer vision, and other machine learning domains~\cite{llama4, llama3, qwen3, deepseek, gpt4}. However, adapting these massive models to downstream tasks often requires fine-tuning millions or even billions of parameters, posing significant challenges for deployment in resource-constrained environments~\cite{peft, prompt, li2021prefix}. Low-Rank Adaptation (LoRA)~\cite{lora} has emerged as a lightweight fine-tuning paradigm, representing weight updates as a product of two much smaller matrices. While LoRA drastically reduces the number of trainable parameters, the underlying dense structure of the original weights still incurs substantial memory and computation overhead~\cite{qlora, lorafa}.

Model pruning offers a complementary approach by eliminating redundant weights to yield sparse networks with reduced storage and faster inference~\cite{hansong-prune, sparsegpt, wanda, compressor}. Ideally, magnitude-based pruning applied to a LoRA-fine-tuned model should combine the benefits of sparsity and low-rank adaptation. However, pruning can disrupt the carefully learned low-rank subspace, and existing dynamic masking strategies often lead to significant performance degradation~\cite{losa}. To guide effective pruning in the LoRA setting, we develop a unified theoretical framework that quantifies the mean-squared error (MSE) induced by different pruning schemes. Our analysis reveals that a static mask applied solely to the frozen base weights achieves the lowest error bound among both static and dynamic approaches, laying the foundation for our Sparsity-Aware Low-Rank Representation fine-tuning (SALR) method.

\begin{figure}[t]
\centering
\includegraphics[width=0.95\columnwidth]{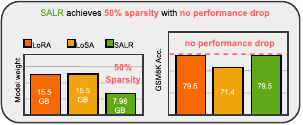}
\caption{Memory-accuracy trade-off on the GSM8K~\cite{gsm8k} benchmark for Llama3-8B~\cite{llama3}, fine-tuned on MetaMath~\cite{metamath}. At 50\% sparsity, SALR maintains the dense LoRA~\cite{lora} baseline accuracy (79.5\%) while reducing model size from 15.5 GB to 7.98 GB. In contrast, LoSA~\cite{losa} at 50\% sparsity suffers a drop to 71.4\% accuracy.}
\label{figure:preview}
\end{figure}

Although static pruning of the base weights minimizes theoretical error, it discards valuable information residing in the pruned elements. To address this loss, SALR introduces a sparsity preservation pruning strategy: rather than permanently zeroing out pruned entries, we capture their residual in an auxiliary low-rank adapter. By applying a truncated singular value decomposition (SVD) to the sparse residual, we have a compact rank-$r$ correction that retains essential information without inflating the parameter count. We further show that this low-rank correction provably reduces the per-entry MSE by a factor of $\bigl(1 - \tfrac{r}{\min(d,k)}\bigr)$, where $d$ and $k$ are the input and output dimensions. However, in this setting, SALR requires the deployment of two or more low-rank adapters, one corresponding to the LoRA fine-tuning module and others serving as sparsity preservation modules. Performing the small matrix multiplications for each adapter sequentially results in under-utilization of modern hardware accelerators. To address this inefficiency, SALR adopts an adapter concatenation scheme: by stacking the low-rank matrices along their rank dimension, all adapter updates are consolidated into a single, larger GEMM (general matrix-matrix multiplication) operation on a shared input. This fusion reduces kernel-launch overhead and maximizes computational throughput when multiple adapters operate on the same input vector.

In recent LoRA-based pruning studies, although most works claim to achieve a specified sparsity ratio and the associated speedup, almost none address the challenge of actual model size compression. Thus, the achieved sparsity does not necessarily translate into reduced model size~\cite{losa, sparselora, deepsparse2, deepsparse1}. This limitation is a major reason why sparse methods lag behind quantization in model compression~\cite{spinfer, flashllm}. Therefore, to enable effective model compression via pruning, SALR integrates a bitmap-based encoding for the pruned base weights along with a two-stage pipelined design. By employing bitmap encoding, SALR efficiently stores the dense elements of sparse weights and enables high-performance bitmap decoding. The two-stage pipeline further alleviates decoding overhead: in the first stage, byte-level masks and lookup tables reconstruct sparse submatrices efficiently, while the LoRA module participates in GEMM computation; concurrently, the second stage feeds these reconstructed blocks into high-performance GEMM kernels. In this manner, the two-stage pipeline sustains compute-bound density throughout all computation phases.

In summary, our contributions are:
\begin{itemize}
    \item We develop a unified, MSE-based theoretical framework for pruning in LoRA-fine-tuned models and prove that applying a static mask to the frozen base weights yields the lowest error bound.
    \item We introduce SALR, a sparsity-preservation pruning method that captures pruned-weight residuals via a truncated-SVD low-rank adapter and provably reduces per-entry MSE.
    \item We propose an adapter concatenation scheme that stacks all low-rank adapters into a single GEMM operation, minimizing kernel-launch overhead and maximizing hardware accelerator utilization.
    \item We design a practical compression and deployment pipeline using bitmap encoding for actual model-size reduction and a two-stage, pipelined decoding+GEMM approach to sustain compute-bound throughput.
\end{itemize}

\section{Preliminary}\label{sec:preliminary}
In this section, we present the essential background and theoretical underpinnings necessary to understand the challenges and methodologies associated with Sparsity-Aware Low-Rank Representation fine-tuning (SALR).

Let the weight matrix be denoted as $W\in\mathbb{R}^{d\times k}$, where $d$ represents the number of input features and $k$ denotes the number of output classes. The magnitude-based pruning operation can be formulated as
$$
\hat{W}_{ij} =
\begin{cases}
W_{ij}, & |W_{ij}| > T_p, \\
0, & |W_{ij}| \leq T_p,
\end{cases}
$$
where $T_p$ is the pruning threshold determined by the desired sparsity ratio $p$. In this manner, pruning can significantly reduce the memory footprint and computational cost of the model, thereby enabling the deployment of large-scale models in resource-constrained environments. In the following, we analyze the error introduced by pruning and discuss how it can be mitigated through low-rank adaptation techniques.

\begin{theorem}\label{theorem:1}
Let $W\sim\mathcal{N}(0,\sigma^2)$ and for a given pruning ratio $p\in[0,1)$ we choose the threshold $T_p$ such that
$$
P\bigl(|W|\le T_p\bigr)=p,
\quad\Longrightarrow\quad
T_p = \sigma\Phi^{-1}\Bigl(\tfrac{1+p}2\Bigr).
$$
Then the mean-squared error (MSE) of this pruning is
$$
\mathrm{MSE}(p)
=
\mathbb E\bigl[(W-\hat W)^2\bigr]
=
2\sigma^2\Bigl[\Phi(t_p)-\tfrac12 - t_p\varphi(t_p)\Bigr],
$$
where
$t_p = \Phi^{-1}\bigl(\tfrac{1+p}2\bigr)$
and
$\varphi(t)=\tfrac1{\sqrt{2\pi}}e^{-t^2/2}$
is the standard normal PDF.
\end{theorem}

Based on the theorem above, if we prune $50\%$ of the entries in the weight matrix, we have
$$
t_{0.5}
=\Phi^{-1}\Bigl(\tfrac{1+0.5}2\Bigr)
=\Phi^{-1}(0.75)
\approx0.674,
$$
and the MSE is given by
$$
\mathrm{MSE}(0.5)
=2\sigma^2\Bigl[\Phi(0.674)-\tfrac12
-0.674\varphi(0.674)\Bigr].
$$
Since $\Phi(0.674)=0.75$ and
$\varphi(0.674)\approx0.318$, we find
\begin{align*}
\mathrm{MSE}(0.5)
&=2\sigma^2\bigl[0.75-0.5 -0.674\cdot0.318\bigr]\\
&\approx2\sigma^2(0.25-0.214)
\approx0.072\sigma^2.
\end{align*}

This result demonstrates that pruning a substantial proportion of model parameters leads to only a modest increase in MSE, thereby establishing the foundational performance guarantees for applying pruning as an effective strategy for large model compression.

However, the pruning operation in LoRA fine-tuning differs in several respects. Due to the introduction of LoRA, the model parameters are represented as a linear combination of low-rank matrices and the original weights, i.e., $W = W_0 + AB$, which implies that pruning may have to account for the effects of the low-rank structure. Specifically, the pruning procedure should aim to reduce the number of parameters while preserving the integrity of the low-rank structure as much as possible. A typical approach is to prune both $W$ and the subspaces of the low-rank matrices $A$ and $B$~\cite{losa}. This strategy ensures that the sparse, low-rank adapter is effectively integrated into the pruned model without altering the model's sparsity after training. However, because pruning is applied to the low-rank subspace, which is designed to compensate for the loss of parameters in the original model, this approach often results in significant performance degradation compared to LoRA.

In addition to pruning both $A$ and $B$ simultaneously, a simpler alternative is to prune only $W_0$, that is, the frozen original weights. The advantage of this approach is that it minimizes the impact of pruning on the low-rank matrices $A$ and $B$, thereby better preserving the low-rank structure of the model. Furthermore, the above methods can be combined, such as deriving a pruning mask from $AB$ and subsequently applying it to $W_0$, which allows for compression of the base model weights while taking the low-rank structure into consideration.

Consider a single pruning step during LoRA fine-tuning, where the objective is to minimize the distance between $W$ and $\hat W$, with $W$ and $\hat W$ denoting the weights before and after pruning, respectively. In the context of LoRA fine-tuning, $W = W_0 + AB$, where $W_0$ is the frozen original weight matrix and $A$ and $B$ are low-rank matrices.
Given the objective, we derive the pruning error bound for such various methods as follows.

\begin{theorem}\label{theorem:2}
Let $W_{0,ij}\sim\mathcal N(0,\sigma^2)$ and $\Delta_{ij}=(A^*B^*)_{ij}\sim\mathcal N(0,\tau^2)$ be independent, where $A^*$ and $B^*$ be the optimal values of $A$ and $B$; $U_{ij}=W_{0,ij}+\Delta_{ij}\sim\mathcal N(0,V^2)$ with $V^2=\sigma^2+\tau^2$; the global prune rate is $p\in[0,1)$, and define the prune operation as the same as above. Let $E_1(p),E_2(p),E_3(p)$ be the per-entry MSEs of

\begin{itemize}
    \item \textbf{Method 1 (static mask on $W_0$)}: prune the smallest entries of $|W_0|$ at rate $p$.
    \item \textbf{Method 2 (dynamic mask driven by $U$, but prune only $W_0$)}.
    \item \textbf{Method 3 (dynamic mask on the full $U=W_0+\Delta$)}.
\end{itemize}

Then the MSEs are
\begin{align}
E_1(p)
&=2\sigma^2\mathcal{Q}(t_p),\\
E_2(p)
&=\frac{\sigma^2\tau^2}{\sigma^2+\tau^2}p
+2\frac{\sigma^4}{\sigma^2+\tau^2}
\mathcal{Q}(t_p)\\\notag
E_3(p)
&=2(\sigma^2+\tau^2)\mathcal{Q}(t_p)\notag\\
\end{align}
where $\mathcal{Q}(t)$ is defined as
$$
\mathcal{Q}(t) \coloneqq \Phi(t) - \tfrac{1}{2} - t\varphi(t).
$$

Moreover, for every $p$,
$$
E_1(p)\le E_3(p)\le E_2(p).
$$

\end{theorem}

\subsection{System Efficiency Limitations in Existing Work}

Table \ref{table:method-compare} compares the latest pruning methods leveraging low-rank adaptation to ours. LoSA and SparseLoRA diverge significantly in both their algorithms and system-level design. LoSA utilizes the third pruning approach described above: it simultaneously prunes both $W$ and the low-rank modules, ensuring the final merged model maintains global sparsity. However, as indicated by our earlier analysis, LoSA's error bounds are relatively high, resulting in only marginal performance retention compared to the original, unpruned model.

In contrast, SparseLoRA uses a dynamic pruning scheme, selecting which elements of $W$ to compute in response to each input during training. This leads to faster fine-tuning, and by not pruning the low-rank subspace, its accuracy matches that of standard LoRA. However, SparseLoRA does not actually prune the deployed model: at inference time, the model remains dense and thus offers no speedup or compression benefits during deployment, i.e., the improvements exist only during training.

\begin{table}
\centering
\addtolength{\tabcolsep}{-3pt}
\small
\begin{tabular}{l|ccc} 
\hline
Method & Performance & Model & Speedup \\ 
\hline
LoSA (ICLR2025) & Low & \textbf{Sparse} & \textbf{Y} \\
SparseLoRA (ICML2025) & \textbf{High} & Dense & N \\
SALR (ours) & \textbf{High} & \textbf{Sparse} & \textbf{Y} \\
\hline
\end{tabular}
\caption{Comparison of LoSA, SparseLoRA and SALR across three criteria: \emph{Performance} indicates end-task accuracy relative to a dense LoRA baseline; \emph{Model} denotes whether the fine-tuned weights remain dense or become sparse; and \emph{Speedup} reports whether true throughput gains for inference are realized. LoSA applies static pruning to achieve sparsity and speedup at the cost of performance; SparseLoRA retains a dense model to match LoRA accuracy but cannot compress or accelerate inference; SALR combines sparse pruning with low-rank residual adapters to deliver both high accuracy and genuine inference speedup.}
\label{table:method-compare}
\end{table}

\section{Methodology}

In this section, we detail the design of our SALR. Based on the error analysis presented in the previous section, we adopt method 1 as our pruning strategy, which achieves the lowest error bound by applying a static mask solely to $W_0$. However, although this approach reduces the error bound, it does not take into account the characteristics of the low-rank structure during pruning. Moreover, it still introduces errors, as pruned elements are typically discarded and their information is lost. To address this limitation, we propose a sparsity preservation pruning strategy, in which the information of the pruned elements is retained. Specifically, during pruning, the pruned elements are stored in an additional low-rank adapter, allowing this information to be leveraged in subsequent training stages. In this manner, we are able to preserve information pertinent to the low-rank structure while pruning, thereby enhancing overall model performance.

\begin{figure*}[t]
\centering
\subfigure[The overall pipeline of SALR]{\label{fig:method-overview}\includegraphics[width=0.7\linewidth]{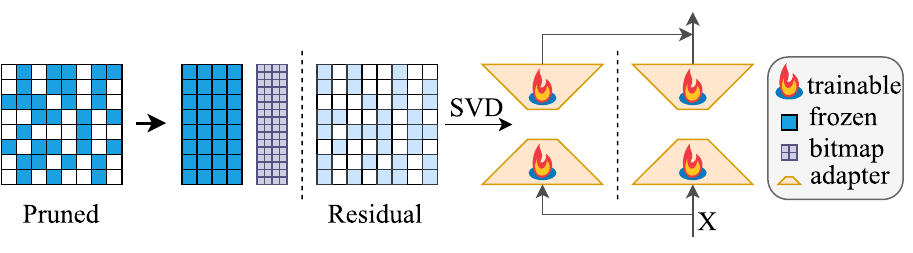}}
\subfigure[Bitmap decoding]{\label{fig:bitmap}\includegraphics[width=0.25\linewidth]{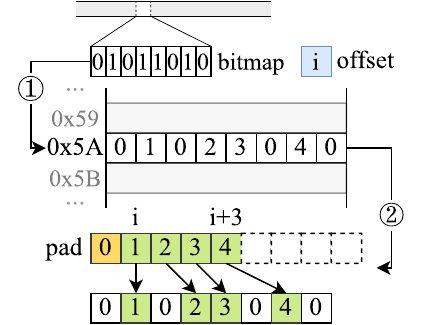}}
\caption{Overview of SALR. (a) SALR first prunes the base model, resulting in a pruned module and a corresponding residual matrix. Only the bitmap and the dense entries of the pruned module are stored. Subsequently, SALR decomposes the residual into a single low-rank pair using the optimal rank-$r$ approximation. (b) Bitmap decoding of sparse weights.}
\label{figure:overview}
\end{figure*}

\subsection{Sparsity Preservation Pruning}

Decomposing the full-parameter weight matrix $W$ into low-rank pairs is an effective approach to reducing the number of parameters while maintaining model performance\cite{salora, pissa, dora, adalora, olora}. Given the pruned weight matrix $\hat W$, the residual matrix can be defined as $E = W - \hat W$. The objective is to obtain a low-rank approximation of $E$ that preserves essential residual information. This is achieved by applying a truncated singular value decomposition (SVD) to $E$ and retaining only the top $r$ singular values. Since the residual matrix $E$ is sparse and often contains negligible information, such decomposition can substantially reduce the parameter count while still capturing the essential information required for model performance. Furthermore, we demonstrate that applying a rank-$r$ low-rank correction to the pruned residual matrix $E$ can effectively bound the mean squared error (MSE) per entry.

\begin{theorem}\label{theorem:3}
Under the same assumptions as Theorem 1, let the residual matrix $E = W - \hat W$. Then
$$
\mathbb{E}\bigl\|E\bigr\|_F^2
= dk\mathrm{MSE}(p).
$$
Now let $E_r$ be the best rank-$r$ approximation to $E$ (i.e.\ its truncated SVD), $q=\min(d,k)$ and the singular values of $E$ as $\sigma_1\ge\dots\ge\sigma_q$, the Eckart-Young theorem~\cite{eckart} gives
$$
\|E - E_r\|_F^2 =\sum_{i=r+1}^q\sigma_i^2.
$$
In the worst case (uniformly distributed spectrum) one has
$\sum_{i=r+1}^q\sigma_i^2\le\frac{q-r}{q}\sum_{i=1}^q\sigma_i^2$.  Taking expectations,
$$
\E\|E - E_r\|_F^2
\le
\Bigl(1-\tfrac{r}{q}\Bigr)\E\|E\|_F^2
=
(dk)\Bigl(1-\tfrac{r}{\min(d,k)}\Bigr)\mathrm{MSE}(p).
$$
Dividing by $dk$ yields the per-entry bound
$$
\mathrm{MSE}_{\rm prune+SVD}(p,r)
\le
\Bigl(1 - \tfrac{r}{\min(d,k)}\Bigr)\mathrm{MSE}(p).
$$
\end{theorem}

According to Theorem \ref{theorem:3}, this SVD residual can closely approximate the pruning error. However, if this residual is fixed during training and only the LoRA adapters $A$ and $B$ are fine-tuned, then the subspaces of $A$ and $B$ must compensate for at least two aspects: \textit{the useful structure that was pruned from $W$}, in addition to \textit{any further task-specific adaptation required}. This burden may be excessive for a small adapter. It is important to note that the subspace of the SVD residual encompasses not only the residual itself; by appropriately adjusting the residual, we can compensate the pruned $\hat W$ with global information without affecting its sparsity. Thus, it is necessary to fine-tune not only $A$ and $B$ to enhance task-specific performance, but also to adjust the SVD residual in order to control task-specific sparsity.
Finally, we derive the optimal learning rate for updating the SVD-residual matrix during fine-tuning, as shown in Theorem~\ref{theorem:4}.

\begin{theorem}\label{theorem:4}
Fix LoRA adapters $A,B$ and let the residual update subproblem be
$$
L(M)=\frac12\|X M - R\|_F^2,
R = Y - X\bigl(\hat W + A B\bigr),
$$
where $X$ and $Y$ be the input and target output respectively, $M$ represents the optimization variable for the residual.

Then $L$ is convex with
$$
\nabla_M L = X^\top(XM - R),
\mathrm{Hess}_M L = I_k \otimes (X^\top X),
$$
and its gradient is Lipschitz with constant
$$
L_{\rm SVD} =\lambda_{\max}(I_k\otimes X^\top X)
=\sigma_{\max}(X)^2.
$$
Hence, gradient descent converges for any step-size
$$
0<\eta<\frac{2}{L_{\rm SVD}}
=\frac{2}{\sigma_{\max}(X)^2},
$$
and the choice minimizing the worst case contraction factor is
$$
\eta^*_{\rm SVD} =\frac{1}{\lambda_{\max}(X^\top X)}
=\frac{1}{\sigma_{\max}(X)^2}.
$$
\end{theorem}

In practice we estimate $\sigma_{\max}(X)$ by a few power-iterations on a representative mini-batch every epoch, and then set $\eta_{\rm SVD}\approx1/\bigl(\sigma_{\max}(X)^2\bigr)$ (or more conservatively, half this value) to guarantee stable and efficient convergence of the SVD-residual updates.

\subsection{Concatenating Multi-LoRA adapters}
By concatenating all adapters into $A_{\rm cat}$ and $B_{\rm cat}$, we replace $2n$ small matrix multiplications with two larger ones, thereby reducing kernel-launch overhead and improving hardware utilization whenever the same $x$ is fed to multiple LoRA adapters.

Specifically, let the input $x\in\mathbb{R}^{d_{\mathrm{in}}}$, and let the $i$-th LoRA Adapter be defined as
$$
A_i \in \mathbb{R}^{d_{\mathrm{in}}\times r},
  B_i \in \mathbb{R}^{r\times d_{\mathrm{out}}}.
$$
The incremental computation for a single adapter is given by $\Delta y_i = xA_iB_i =(xA_i)B_i\in\mathbb{R}^{1\times d_{\mathrm{out}}}.$ If the updates are accumulated sequentially, this requires $2n$ small matrix multiplications. To improve computational efficiency, we define the concatenated matrices as
$$
A_{\mathrm{cat}}
=
\begin{bmatrix}
A_1^\top\\
A_2^\top\\
\vdots\\
A_n^\top
\end{bmatrix}
^{\top}
\in\mathbb{R}^{d_{\mathrm{in}}\times (nr)},
B_{\mathrm{cat}}
=
\begin{bmatrix}
B_1\\
B_2\\
\vdots\\
B_n
\end{bmatrix}
\in\mathbb{R}^{(nr)\times d_{\mathrm{out}}}.
$$
With this construction, only two matrix multiplications are required to obtain the cumulative update from all adapters:
$$
\Delta y
=\sum_{i=1}^n (xA_i)B_i
\in\mathbb{R}^{1\times d_{\mathrm{out}}}.
$$
The final output is then given by $y = xW + \Delta y.$

\subsection{Mapping Sparse Weights and Pipeline Design}
Traditional CSR-format sparse representations incur significant indexing overhead. In contrast, bitmap encoding largely eliminates this cost, though it requires an efficient decoding mechanism to avoid throughput overhead. We show that this decoding process can be highly parallelized.

\textbf{Mapping Sparse Weights.} We define the bitmap as
$$
B\in\{0,1\}^{d_{\mathrm{in}}\times d_{\mathrm{out}}},
\quad
B_{ij}=\begin{cases}
1,&\hat W_{ij}\neq0,\\
0,&\hat W_{ij}=0.
\end{cases}
$$
All nonzero elements are stored in a compact array $v\in\mathbb{R}^{\mathrm{nnz}(\hat W)}$ in row-major order, where $\mathrm{nnz}(\hat W)=\sum_{i,j}B_{ij}$.

To accelerate sparse matrix reconstruction, we partition each row into several byte blocks, with every 8 columns forming one group. For the $b$-th byte block of row $i$, we define the bitmask as
$\mathit{mask}_{i,b}
=\sum_{t=0}^{7}B_{i,8b+t}2^t
\in\{0,\dots,255\}.$
Let $\mathrm{bit}_t(m)$ denote the $t$-th binary digit of integer $m$:
$$
\mathrm{bit}_t(m)
=\bigl\lfloor (m\div2^t)\bigr\rfloor\bmod2,\quad t=0,\dots,7,
$$
then
$\mathit{mask}_{i,b}
=\sum_{t=0}^7\mathrm{bit}_t\bigl(B_{i,8b+t}\bigr)2^t.$
We further define the popcount function as
\begin{align*}
\mathrm{popcount}(m)
&=\sum_{t=0}^7\mathrm{bit}_t(m)\\
&=\#\bigl\{t:0\le t\le7,\mathrm{bit}_t(m)=1\bigr\},
\end{align*}
which returns the number of 1's in the binary representation of $m$. Thus, the number of nonzero elements in the $b$-th byte block of row $i$ is
$$
k_{i,b}=\mathrm{popcount}(\mathit{mask}_{i,b}),
$$
corresponding to a segment in the compact array, $v_{i,b}=(v_{i,b}^0,v_{i,b}^1,\dots,v_{i,b}^{k_{i,b}-1}).$

We precompute a lookup table $\mathrm{LUT}:\{0,\dots,255\}\to\{-1,0,1,\dots,7\}^8,$ such that for any $\mathit{mask}$, if its $t$-th bit is 1, then $\mathrm{LUT}(\mathit{mask})_t$ gives the corresponding index in the nonzero segment $v_{i,b}$; otherwise, it is $-1$. Let
$$
\mathrm{LUT}(\mathit{mask})=(\ell_0,\dots,\ell_7),
$$
then the sparse reconstruction rule can be expressed as
$$
\forall t=0,\dots,7:\quad
\hat W_{i,8b+t}
=\begin{cases}
v_{i,b}^{\ell_t},&\ell_t\ge0,\\
0,&\ell_t=-1.
\end{cases}
$$

\textbf{Pipeline Design.}  
To fully exploit hardware capabilities, we decouple sparse bitmap decoding from GEMM computation and construct a two-stage pipeline. During the decoding stage, CUDA cores sequentially read the bitmap $B$ and compact array $v$ by byte block, and reconstruct contiguous sparse submatrix blocks using the precomputed lookup table $\mathrm{LUT}$. In the computation stage, the recovered submatrix blocks are processed by invoking Tensor Core to perform dense-dense multiplication. The two stages are connected via a ring buffer: while the decoding stage processes block $b+1$, the computation stage can simultaneously execute the multiplication for block $b$. This design eliminates blocking due to sparse format conversion during the main computation, maximizes hardware throughput, and significantly reduces memory access and scheduling overhead.

\section{Experimental Results}

\begin{table*}[t]
\begin{center}
\addtolength{\tabcolsep}{-3pt}
\small
\begin{tabular}{l|ccc|ccc|ccc} 
\hline
Model & \multicolumn{3}{c|}{Llama2-7B} & \multicolumn{3}{c|}{Llama3-8B} & \multicolumn{3}{c}{Mixtral-8x7B} \\
Dataset & MMLU & GSM8K & Sparsity & MMLU & GSM8K & Sparsity & MMLU & GSM8K & Sparsity \\ 
\hline
Pretrained & 45.6 & 35.5 & - & 66.0 & 72.0 & - & 70.6 & 58.4 & - \\
LoRA & 56.0 & 56.8 & - & 69.2 & 79.5 & - & 71.0 & 79.2 & - \\
\midrule
LoSA \cite{losa} & 45.0 & 34.2 & 50\% & 64.4 & 71.4 & 50\% & 69.2 & 57.9 & 50\% \\
SparseLoRA \cite{sparselora} & \bd{56.0} & 37.6 & - & \bd{69.0} & 72.0 & - & 70.9 & 78.0 & - \\
DeepSparse \cite{deepsparse2} & 45.1 & 36.5 & 50\% & 60.4 & 47.9 & 50\% & 70.0 & 59.0 & 50\% \\
Ours & \bd{56.0} & \bd{56.7} & 50\% & 68.2 & \bd{79.5} & 50\% & \bd{71.4} & \bd{79.1} & 50\% \\
\hline
\end{tabular}
\end{center}
\caption{Performance comparison on the MMLU and GSM8K benchmarks. We utilize a global $50\%$ sparsity for all methods except Pretrained and LoRA. The rank is set to 64. "-" denotes N/A.}
\label{tab:performance}
\end{table*}

We conduct extensive experiments to evaluate the effectiveness of our SALR across a range of benchmarks and models. First, we focus on supervised fine-tuning (SFT) of state-of-the-art LLMs in the MATH and multidisciplinary domains, comparing the performance of SALR against its competitors, i.e., LoRA-based pruning methods, on MMLU and GSM8K benchmarks.
Additionally, we examine the system efficiency of various methods during fine-tuning, demonstrating that SALR significantly reduces model size while maintaining fine-tuning throughput. The models, datasets, metrics, and baselines used in our experiments are detailed below, with further experimental settings provided in supplementary due to space constraints.

\textbf{Models.} We fine-tune a diverse selection of LLMs, including Llama2-7B~\cite{llama2}, Llama3-8B~\cite{llama3}, Mixtral-8x7B~\cite{mixtral}, DeepSeek-V2-Lite~\cite{deepseek}.

\textbf{Datasets.}~In alignment with prior studies~\cite{loraga, lorapro}, our experiments span a variety of datasets tailored to specific task types. We utilize MetaMath~\cite{metamath} for the MATH domain,  auxiliary multiple-choice training questions from ARC, MC-TEST, OBQA, RACE, etc., for the multidisciplinary domain.

\textbf{Evaluation Metrics.}~Following the evaluation protocols of prior works such as QLoRA~\cite{qlora}, LLM-Adapters~\cite{llmadapters}, and LoRA-Pro~\cite{lorapro}, we primarily assess the zero-shot performance of fine-tuned LLMs across various benchmarks. for MMLU we report 5-shot accuracy, and for GSM8K we report zero-shot accuracy.

\textbf{Baselines.}~We compare the performance of SALR against pretrained baseline, the standard LoRA, and several recent LoRA-based pruning methods, including LoSA~\cite{losa}, SparseLoRA~\cite{sparselora}, DeepSparse~\cite{deepsparse1}. Please refer to supplementary for detailed introduction of baselines.

\subsection{Performance on Domain-Specific Tasks}\label{sec:domain}
Across all evaluated models and benchmarks, our method consistently outperforms traditional pruning techniques such as LoSA, SparseLoRA, and DeepSparse. For instance, on Llama2-7B, SALR achieves 56.0 on MMLU and 56.7 on GSM8K, which is notably higher than LoSA (45.0/34.2), SparseLoRA (56.0/37.6), and DeepSparse (45.1/36.5). Similarly, for Llama3-8B, our approach yields 68.2 on MMLU and 79.5 on GSM8K, closely aligning with the performance of LoRA (69.2/79.5), and surpassing other pruning baselines by a significant margin.

Notably, while LoRA remains a strong baseline for parameter-efficient fine-tuning, our method reaches comparable or even superior performance, particularly on Mixtral-8x7B, where SALR scores 71.4 on MMLU and 79.1 on GSM8K, effectively matching or exceeding the LoRA results while providing significant model compression.
These results highlight the capability of our approach to maintain high performance while substantially reducing model size. The alignment with LoRA on key benchmarks further demonstrates that our pruning technique does not compromise the model's ability to generalize across diverse tasks.

\begin{figure}[t]
\centering
\subfigure[LoSA]{\includegraphics[width=0.45\columnwidth]{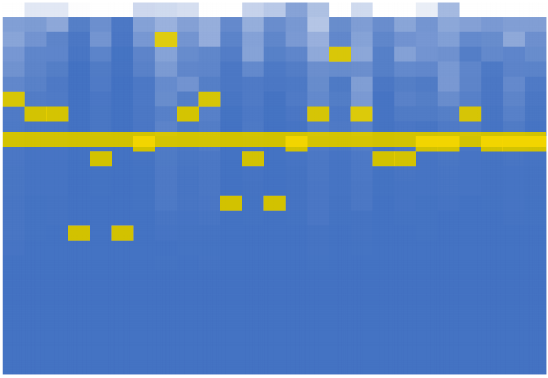}}
\subfigure[SALR]{\includegraphics[width=0.45\columnwidth]{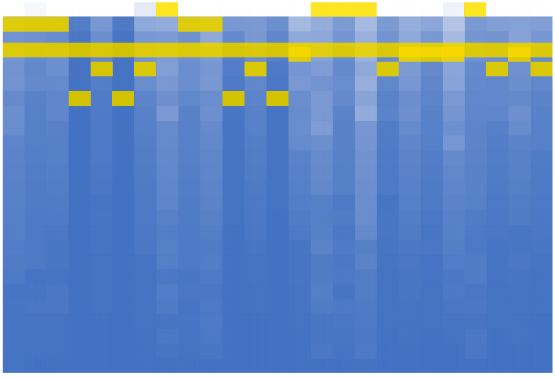}}
\caption{Normalized cumulative singular-value energy spectra of the residual correction matrices for LoSA and SALR on Llama3-8B after fine-tuning on MetaMath.}
\label{figure:svd}
\end{figure}

Figure~\ref{figure:svd} shows the normalized singular-value energy spectra of the weight updates for LoSA and SALR. Concretely, if $(\sigma_i)_{i=1}^q$ are the singular values of the residual correction matrix $E$, we show
$$
\frac{\sum_{j=1}^i\sigma_j^2}{\sum_{j=1}^q\sigma_j^2}
\quad\text{where}\quad
i=1,2,\dots,q,
$$
and we mark the cube (with yellow) with the smallest index $i_{0.99}$ such that
$\sum_{j=1}^{i_{0.99}}\sigma_j^2 / \sum_{j=1}^q\sigma_j^2 \ge 0.99$.  
$i_{0.99}^{\rm LoSA}\ll i_{0.99}^{\rm SALR}$ is reported, indicating LoSA requires far fewer singular values to reach 99\% energy, whereas SALR retains a much larger tail of the spectrum.  In other words, SALR preserves substantially more of the residual's singular-value energy, resulting in a denser effective update.
This finding directly corroborates the Theorem~\ref{theorem:3}. By retaining a larger number of singular values ($r$), SALR drives down the RHS of this bound, preserving more of the original update energy and delivering higher accuracy.

\subsection{System Efficiency}
\textbf{Model Compression and Throughput in Fine-Tuning.}
Table~\ref{table:system-efficiency} reports the fine-tuning memory footprint and sustained throughput for LoSA and SALR. LoSA consumes 27.1 GB of GPU memory and achieves 74.5 TFLOPS, whereas SALR reduces the memory to 19.2 GB and boosts throughput to 89.2 TFLOPS. Remarkably, given sparsity at 50\%, our method achieves a 2x reduction in model size for Llama3-8B which matches with LoSA, demonstrating the effectiveness of our bitmap decoding strategy in compressing large language models. 
The key inefficiency in LoSA arises from its two-stage dense update: $\Delta W = AB$ and then $\Delta Y = X\Delta W.$ They are two compute-intensive GEMM operations and contribute to the activation update after applying the sparsity mask.
In contrast, since SALR prunes only W, thus it can align the computation with LoRA so that only low-rank products are formed: $U = XA, \Delta Y = UB,$
with $U\in\mathbb R^{N\times r}$ and $r\ll\min(d_{\rm in},d_{\rm out})$.  This strategy replaces the expensive full-rank GEMM by two smaller GEMMs of complexity $O(Nd_{\rm in}r)$ and $O(Nrd_{\rm out})$.
Moreover, by concatenating all adapters along the rank dimension, SALR fuses multiple updates into a single GEMM, further reducing kernel-launch overhead.
Overall, SALR's fused low-rank adapter kernel and sparsity decoding yield a $\sim30\%$ reduction in fine-tuning memory and a $\sim20\%$ increase in TFLOPS relative to LoSA.
\begin{table}
\small
\centering
\addtolength{\tabcolsep}{-3pt}
\begin{tabular}{l|cccc} 
\hline
Method & Sparsity & FT mem (GB)$\downarrow$ & FT TFLOPS$\uparrow$ & \# Comp \\ 
\hline
LoRA & - & 26.7 & 91.9 & - \\
LoSA & 50\% & 27.1 & 74.5 & 2.0 x \\
SALR & 50\% & 19.2 & 89.2 & 2.0 x \\
\hline
\end{tabular}
\caption{Fine-tuning GPU memory footprint and sustained throughput (in TFLOPS) on Llama3-8B for LoRA (dense), LoSA (50\% sparsity) and SALR (50\% sparsity). \# Comp denotes model compression rate.}
\label{table:system-efficiency}
\end{table}

\begin{table}
\small
\centering
\addtolength{\tabcolsep}{-2pt}
\begin{tabular}{lccc} 
\hline
Method (Sparsity) & GSM8K$\uparrow$ & \makecell{Throughput\\(tokens/s)}$\uparrow$ & Speedup \\ 
\hline
LoRA (N/A) & 79.5 & 60.1 & 1.0 x \\
SparseLoRA (N/A) & 72 & 60.1 & 1.0 x \\
LoSA (2:4) & 69.4 & 113.5 & 1.9 x \\
SALR (2:4) & 78.9 & 104.9 & 1.7 x \\
\hline
\end{tabular}
\caption{GSM8K accuracy, inference throughput (tokens/s) and inference speedup of Llama3-8B. Dense LoRA and SparseLoRA (no sparsity) serve as baselines, LoSA and SALR are fine-tuned and evaluated under 2:4 semi-structured pattern. Results are collected on 1x RTX4090.}
\label{table:inference}
\end{table}

\textbf{Inference Speedup.} To assess the end-to-end benefits of SALR at inference time, we follow the N:M sparsity protocol of~\cite{losa} and measure the throughput and speedup on GSM8K of Llama3-8B. We include dense LoRA and SparseLoRA, which incur no model compression, as well as LoSA and SALR under 2:4 sparsity.  Results are shown in Table~\ref{table:inference}.
Despite operating at 50\% sparsity, SALR retains nearly the same GSM8K accuracy as dense LoRA while delivering a 1.7$\times$ speedup.  Compared to LoSA, which suffers a larger accuracy drop under 2:4 sparsity, SALR achieves significantly higher end-task performance (78.9\% vs.\ 69.4\%) at only a modest reduction in throughput.

\subsection{Ablation Study}

\textbf{Train Residual to Boost Performance.} 
To quantify the impact of updating the sparsity-preservation residual during fine-tuning, we compare three variants: standard LoRA, SALR with the residual kept frozen, and SALR with the residual trainable. Results on Llama2-7B and Llama3-8B are summarized in Table~\ref{table:ablation-residual}.
When the SVD-derived residual is held fixed, SALR suffers a notable accuracy drop (-1.8 on Llama2-7B, -2.4 on Llama3-8B), indicating that the pruned-weight correction learned from the pretrained model is not fully aligned with the downstream task. By making the residual trainable alongside the LoRA adapters, SALR recovers nearly all of this lost performance, matching LoRA on Llama2-7B and reducing the gap to just 1.0 on Llama3-8B. This demonstrates that fine-tuning the sparsity-preservation residual is crucial for adapting the preserved information to the task.

\textbf{Coupled with Quantization.} To further shrink the memory footprint of SALR for ultra-large models, we combine a 20\% static sparsity mask with NF4 quantization, yielding Quantized SALR (QSALR).  Table~\ref{table:qsalr} reports end-task accuracy and peak memory for DeepSeek-V2 and Mixtral-8x7B. By applying NF4 quantization on top of 20\% sparse SALR, QSALR achieves a $\sim$5× reduction in model size for both models, dropping from 31.8 GB to 6.5 GB on DeepSeek-V2 and from 93.9 GB to 19.2 GB on Mixtral-8x7B.  The performance degradation is minimal (-0.6 on DeepSeek-V2 and none on Mixtral-8x7B), confirming that the structured sparsity and low-precision quantization are highly complementary. For the Huawei NPU deployment of Mixtral-8x7B, QSALR demonstrates similar efficiency and accuracy as on GPU, showing that the quantized sparse representation generalizes well across heterogeneous hardware. This combination enables efficient deployment of very large LLMs in constrained environments without sacrificing task performance.

\begin{table}
\small
\centering
\begin{tabular}{l|cc} 
\hline
\multirow{2}{*}{Method} & \multicolumn{2}{c}{MMLU$\uparrow$} \\
 & Llama2-7B & Llama3-8B \\ 
\hline
LoRA & 56 & 69.2 \\
SALR w/ frozen residual & 54.2 & 66.8 \\
SALR w/ trainable residual & 56 & 68.2 \\
\hline
\end{tabular}
\caption{Ablation of residual update on MMLU accuracy.}
\label{table:ablation-residual}
\end{table}

\begin{table}
\small
\centering
\addtolength{\tabcolsep}{-5.5pt}
\begin{tabular}{l|cc|cc|cc} 
\hline
\multirow{2}{*}{Model} & \multicolumn{2}{c|}{DeepSeek-V2-Lite} & \multicolumn{2}{c|}{Mixtral-8x7B} & \multicolumn{2}{l}{Mixtral-8x7B(NPU)} \\
 & Acc.$\uparrow$ & Size (GB)$\downarrow$ & Acc.$\uparrow$ & Size (GB)$\downarrow$ & Acc.$\uparrow$ & Size (GB)$\downarrow$ \\ 
\hline
LoRA & 71 & 31.8 & 79.2 & 93.9 & 79.2 & 94.0 \\
QSALR & 70.4 & 6.5 & 79.2 & 19.2 & 78.0 & 19.2 \\
\hline
\end{tabular}
\caption{GSM8K accuracy and model size of LoRA and SALR with quantization (20\% sparsity + NF4).}
\label{table:qsalr}
\end{table}

\begin{table}[ht]
\small
\centering
\begin{tabular}{l|c} 
\hline
Method (Sparsity) & Llama3-8B GSM8K$\uparrow$ \\ 
\hline
LoRA (N/A) & 79.5 \\
SALR~(10\%) & 79.5 \\
SALR~(30\%) & 80.1 \\
SALR~(50\%) & 79.5 \\
\hline
\end{tabular}
\caption{Effect of sparsity level on GSM8K accuracy.}
\label{table:ablation-sparsity}
\end{table}

\textbf{Sparsity-Accuracy Trade-off.} We evaluate how varying the global sparsity level affects SALR's performance. Table~\ref{table:ablation-sparsity} reports GSM8K accuracy on Llama3-8B for sparsity levels from 10\% to 50\%.
Across all sparsity levels up to 50\%, SALR maintains accuracy nearly identical to dense LoRA. Notably, the 30\%-sparse model even slightly surpasses the baseline (80.1\% vs.\ 79.5\%), suggesting that moderate sparsity can act as a regularizer. Furthermore, SALR can aggressively prune up to half of the weights without compromising the accuracy.

\section{Conclusion}

We have presented SALR, a novel fine-tuning paradigm that unifies magnitude-based pruning with low-rank adaptation under a principled MSE framework. Our analysis shows that statically pruning only the frozen base weights minimizes the pruning error bound, and that recovering the discarded residual via a truncated-SVD low-rank adapter provably reduces per-entry MSE.  To maximize hardware efficiency, we fuse multiple adapters into a single concatenated GEMM and employ bitmap encoding with a two-stage decoding + GEMM pipeline, yielding true model compression and inference speedup.  Empirically, SALR achieves 50\% sparsity on various LLMs with no loss in GSM8K and MMLU accuracy, halves model size, and delivers up to 1.7× inference speedup. Our work paves the way for resource-efficient deployment of large language models in constrained environments.

\section{Acknowledgments}
This work was partially supported by the Guangzhou Municipal Joint Funding Project with Universities and Enterprises under Grant No. 2024A03J0616, and Hong Kong CRF grants under Grant C6015-23G.

\bibliography{cite}

@Misc{peft,
  title =        {PEFT: State-of-the-art Parameter-Efficient Fine-Tuning methods},
  author =       {Sourab Mangrulkar and Sylvain Gugger and Lysandre Debut and Younes Belkada and Sayak Paul},
  howpublished = {\url{https://github.com/huggingface/peft}},
  year =         {2022}
}

@inproceedings{li2021prefix,
  title={Prefix-Tuning: Optimizing Continuous Prompts for Generation},
  author={Li, Xiang Lisa and Liang, Percy},
  booktitle={Proceedings of the 59th Annual Meeting of the Association for Computational Linguistics and the 11th International Joint Conference on Natural Language Processing (Volume 1: Long Papers)},
  pages={4582--4597},
  year={2021}
}

@inproceedings{
lora,
title={Lo{RA}: Low-Rank Adaptation of Large Language Models},
author={Edward J Hu and Yelong Shen and Phillip Wallis and Zeyuan Allen-Zhu and Yuanzhi Li and Shean Wang and Lu Wang and Weizhu Chen},
booktitle={International Conference on Learning Representations},
year={2022},
url={https://openreview.net/forum?id=nZeVKeeFYf9}
}

@misc{llama2,
      title={Llama 2: Open Foundation and Fine-Tuned Chat Models}, 
      author={Hugo Touvron and Louis Martin and Kevin Stone and Peter Albert and Amjad Almahairi and Yasmine Babaei and Nikolay Bashlykov and Soumya Batra and Prajjwal Bhargava and Shruti Bhosale et al.},
      year={2023},
      eprint={2307.09288},
      archivePrefix={arXiv},
      primaryClass={cs.CL}
}

@misc{gpt4,
      title={GPT-4 Technical Report}, 
      author={OpenAI},
      year={2023},
      eprint={2303.08774},
      archivePrefix={arXiv},
      primaryClass={cs.CL}
}

@article{
qlora,
title={QLoRA: Efficient Finetuning of Quantized LLMs},
author={Dettmers, Tim and Pagnoni, Artidoro and Holtzman, Ari and Zettlemoyer, Luke},
journal={arXiv preprint arXiv:2305.14314},
year={2023}
}

@article{llmadapters,
  title={LLM-Adapters: An Adapter Family for Parameter-Efficient Fine-Tuning of Large Language Models},
  author={Hu, Zhiqiang and Lan, Yihuai and Wang, Lei and Xu, Wanyu and Lim, Ee-Peng and Lee, Roy Ka-Wei and Bing, Lidong and Poria, Soujanya},
  journal={arXiv preprint arXiv:2304.01933},
  year={2023}
}

@misc{llama3,
      title={The Llama 3 Herd of Models}, 
      author={Abhimanyu Dubey and Abhinav Jauhri et al.},
      year={2024},
      eprint={2407.21783},
      archivePrefix={arXiv},
      primaryClass={cs.AI},
      url={https://arxiv.org/abs/2407.21783}, 
}

@article{gsm8k,
  title={Training Verifiers to Solve Math Word Problems},
  author={Cobbe, Karl and Kosaraju, Vineet and Bavarian, Mohammad and Chen, Mark and Jun, Heewoo and Kaiser, Lukasz and Plappert, Matthias and Tworek, Jerry and Hilton, Jacob and Nakano, Reiichiro and Hesse, Christopher and Schulman, John},
  journal={arXiv preprint arXiv:2110.14168},
  year={2021}
}

@misc{prompt,
      title={The Power of Scale for Parameter-Efficient Prompt Tuning}, 
      author={Brian Lester and Rami Al-Rfou and Noah Constant},
      year={2021},
      eprint={2104.08691},
      archivePrefix={arXiv},
      primaryClass={cs.CL},
      url={https://arxiv.org/abs/2104.08691}, 
}

@misc{olora,
      title={OLoRA: Orthonormal Low-Rank Adaptation of Large Language Models}, 
      author={Kerim Büyükakyüz},
      year={2024},
      eprint={2406.01775},
      archivePrefix={arXiv},
      primaryClass={cs.CL},
      url={https://arxiv.org/abs/2406.01775}, 
}

@misc{pissa,
      title={PiSSA: Principal Singular Values and Singular Vectors Adaptation of Large Language Models}, 
      author={Fanxu Meng and Zhaohui Wang and Muhan Zhang},
      year={2024},
      eprint={2404.02948},
      archivePrefix={arXiv},
      primaryClass={cs.LG},
      url={https://arxiv.org/abs/2404.02948}, 
}

@misc{lorapro,
      title={LoRA-Pro: Are Low-Rank Adapters Properly Optimized?}, 
      author={Zhengbo Wang and Jian Liang and Ran He and Zilei Wang and Tieniu Tan},
      year={2024},
      eprint={2407.18242},
      archivePrefix={arXiv},
      primaryClass={cs.LG},
      url={https://arxiv.org/abs/2407.18242}, 
}

@article{metamath,
  title={MetaMath: Bootstrap Your Own Mathematical Questions for Large Language Models},
  author={Yu, Longhui and Jiang, Weisen and Shi, Han and Yu, Jincheng and Liu, Zhengying and Zhang, Yu and Kwok, James T and Li, Zhenguo and Weller, Adrian and Liu, Weiyang},
  journal={arXiv preprint arXiv:2309.12284},
  year={2023}
}

@misc{loraga,
    title={LoRA-GA: Low-Rank Adaptation with Gradient Approximation},
    author={Shaowen Wang and Linxi Yu and Jian Li},
    year={2024},
    eprint={2407.05000},
    archivePrefix={arXiv},
    primaryClass={cs.LG},
    url={https://arxiv.org/abs/2407.05000},
}

@misc{adalora,
      title={AdaLoRA: Adaptive Budget Allocation for Parameter-Efficient Fine-Tuning}, 
      author={Qingru Zhang and Minshuo Chen and Alexander Bukharin and Nikos Karampatziakis and Pengcheng He and Yu Cheng and Weizhu Chen and Tuo Zhao},
      year={2023},
      eprint={2303.10512},
      archivePrefix={arXiv},
      primaryClass={cs.CL},
      url={https://arxiv.org/abs/2303.10512}, 
}

@article{dora,
  title={DoRA: Weight-Decomposed Low-Rank Adaptation},
  author={Liu, Shih-Yang and Wang, Chien-Yi and Yin, Hongxu and Molchanov, Pavlo and Wang, Yu-Chiang Frank and Cheng, Kwang-Ting and Chen, Min-Hung},
  journal={arXiv preprint arXiv:2402.09353},
  year={2024}
}

@misc{mixtral,
      title={Mixtral of Experts}, 
      author={Albert Q. Jiang and Alexandre Sablayrolles and Antoine Roux and Arthur Mensch and Blanche Savary and Chris Bamford and Devendra Singh Chaplot and Diego de las Casas and Emma Bou Hanna and Florian Bressand and Gianna Lengyel and Guillaume Bour and Guillaume Lample and Lélio Renard Lavaud and Lucile Saulnier and Marie-Anne Lachaux and Pierre Stock and Sandeep Subramanian and Sophia Yang and Szymon Antoniak and Teven Le Scao and Théophile Gervet and Thibaut Lavril and Thomas Wang and Timothée Lacroix and William El Sayed},
      year={2024},
      eprint={2401.04088},
      archivePrefix={arXiv},
      primaryClass={cs.LG},
      url={https://arxiv.org/abs/2401.04088}, 
}

@article{qwen3,
    title={Qwen3 Technical Report}, 
    author={An Yang and Anfeng Li and Baosong Yang and Beichen Zhang and Binyuan Hui and Bo Zheng and Bowen Yu and Chang Gao and Chengen Huang and Chenxu Lv and Chujie Zheng and Dayiheng Liu and Fan Zhou and Fei Huang and Feng Hu and Hao Ge and Haoran Wei and Huan Lin and Jialong Tang and Jian Yang and Jianhong Tu and Jianwei Zhang and Jianxin Yang and Jiaxi Yang and Jing Zhou and Jingren Zhou and Junyang Lin and Kai Dang and Keqin Bao and Kexin Yang and Le Yu and Lianghao Deng and Mei Li and Mingfeng Xue and Mingze Li and Pei Zhang and Peng Wang and Qin Zhu and Rui Men and Ruize Gao and Shixuan Liu and Shuang Luo and Tianhao Li and Tianyi Tang and Wenbiao Yin and Xingzhang Ren and Xinyu Wang and Xinyu Zhang and Xuancheng Ren and Yang Fan and Yang Su and Yichang Zhang and Yinger Zhang and Yu Wan and Yuqiong Liu and Zekun Wang and Zeyu Cui and Zhenru Zhang and Zhipeng Zhou and Zihan Qiu},
    journal = {arXiv preprint arXiv:2505.09388},
    year={2025}
}

@misc{deepseek,
      title={DeepSeek-R1: Incentivizing Reasoning Capability in LLMs via Reinforcement Learning}, 
      author={DeepSeek-AI},
      year={2025},
      eprint={2501.12948},
      archivePrefix={arXiv},
      primaryClass={cs.CL},
      url={https://arxiv.org/abs/2501.12948}, 
}

@article{sparsegpt,
  title={{SparseGPT}: Massive Language Models Can Be Accurately Pruned in One-Shot}, 
  author={Elias Frantar and Dan Alistarh},
  year={2023},
  journal={arXiv preprint arXiv:2301.00774}
}

@misc{wanda,
      title={A Simple and Effective Pruning Approach for Large Language Models}, 
      author={Mingjie Sun and Zhuang Liu and Anna Bair and J. Zico Kolter},
      year={2024},
      eprint={2306.11695},
      archivePrefix={arXiv},
      primaryClass={cs.CL},
      url={https://arxiv.org/abs/2306.11695}, 
}

@misc{llama4,
  title        = {The Llama 4 herd: The beginning of a new era of natively multimodal AI innovation},
  author       = {Meta},
  year         = 2025,
  note         = {\url{https://ai.meta.com/blog/llama-4-multimodal-intelligence/} [Accessed: 2025-04-05]}
}

@misc{sparselora,
      title={SparseLoRA: Accelerating LLM Fine-Tuning with Contextual Sparsity}, 
      author={Samir Khaki and Xiuyu Li and Junxian Guo and Ligeng Zhu and Chenfeng Xu and Konstantinos N. Plataniotis and Amir Yazdanbakhsh and Kurt Keutzer and Song Han and Zhijian Liu},
      year={2025},
      eprint={2506.16500},
      archivePrefix={arXiv},
      primaryClass={cs.LG},
      url={https://arxiv.org/abs/2506.16500}, 
}

@misc{hansong-prune,
      title={Deep Compression: Compressing Deep Neural Networks with Pruning, Trained Quantization and Huffman Coding}, 
      author={Song Han and Huizi Mao and William J. Dally},
      year={2016},
      eprint={1510.00149},
      archivePrefix={arXiv},
      primaryClass={cs.CV},
      url={https://arxiv.org/abs/1510.00149}, 
}

@misc{losa,
      title={Dynamic Low-Rank Sparse Adaptation for Large Language Models}, 
      author={Weizhong Huang and Yuxin Zhang and Xiawu Zheng and Yang Liu and Jing Lin and Yiwu Yao and Rongrong Ji},
      year={2025},
      eprint={2502.14816},
      archivePrefix={arXiv},
      primaryClass={cs.LG},
      url={https://arxiv.org/abs/2502.14816}, 
}

@misc{deepsparse1,
      title={Sparse Fine-tuning for Inference Acceleration of Large Language Models}, 
      author={Eldar Kurtic and Denis Kuznedelev and Elias Frantar and Michael Goin and Dan Alistarh},
      year={2023},
      eprint={2310.06927},
      archivePrefix={arXiv},
      primaryClass={cs.CL},
      url={https://arxiv.org/abs/2310.06927}, 
}

@misc{deepsparse2,
      title={Enabling High-Sparsity Foundational Llama Models with Efficient Pretraining and Deployment}, 
      author={Abhinav Agarwalla and Abhay Gupta and Alexandre Marques and Shubhra Pandit and Michael Goin and Eldar Kurtic and Kevin Leong and Tuan Nguyen and Mahmoud Salem and Dan Alistarh and Sean Lie and Mark Kurtz},
      year={2024},
      eprint={2405.03594},
      archivePrefix={arXiv},
      primaryClass={cs.CL},
      url={https://arxiv.org/abs/2405.03594}, 
}

@misc{compressor,
      title={Compresso: Structured Pruning with Collaborative Prompting Learns Compact Large Language Models}, 
      author={Song Guo and Jiahang Xu and Li Lyna Zhang and Mao Yang},
      year={2023},
      eprint={2310.05015},
      archivePrefix={arXiv},
      primaryClass={cs.AI},
      url={https://arxiv.org/abs/2310.05015}, 
}

@misc{flashllm,
      title={Flash-LLM: Enabling Cost-Effective and Highly-Efficient Large Generative Model Inference with Unstructured Sparsity}, 
      author={Haojun Xia and Zhen Zheng and Yuchao Li and Donglin Zhuang and Zhongzhu Zhou and Xiafei Qiu and Yong Li and Wei Lin and Shuaiwen Leon Song},
      year={2023},
      eprint={2309.10285},
      archivePrefix={arXiv},
      primaryClass={cs.DC},
      url={https://arxiv.org/abs/2309.10285}, 
}

@inproceedings{spinfer,
author = {Fan, Ruibo and Yu, Xiangrui and Dong, Peijie and Li, Zeyu and Gong, Gu and Wang, Qiang and Wang, Wei and Chu, Xiaowen},
title = {SpInfer: Leveraging Low-Level Sparsity for Efficient Large Language Model Inference on GPUs},
year = {2025},
isbn = {9798400711961},
publisher = {Association for Computing Machinery},
address = {New York, NY, USA},
url = {https://doi.org/10.1145/3689031.3717481},
doi = {10.1145/3689031.3717481},
booktitle = {Proceedings of the Twentieth European Conference on Computer Systems},
pages = {243–260},
numpages = {18},
location = {Rotterdam, Netherlands},
series = {EuroSys '25}
}

@misc{salora,
      title={SaLoRA: Safety-Alignment Preserved Low-Rank Adaptation}, 
      author={Mingjie Li and Wai Man Si and Michael Backes and Yang Zhang and Yisen Wang},
      year={2025},
      eprint={2501.01765},
      archivePrefix={arXiv},
      primaryClass={cs.LG},
      url={https://arxiv.org/abs/2501.01765}, 
}

@article{eckart,
  title={The approximation of one matrix by another of lower rank},
  author={Eckart, Carl and Young, Gale},
  journal={Psychometrika},
  volume={1},
  number={3},
  pages={211--218},
  year={1936},
  publisher={Springer-Verlag}
}

@misc{lorafa,
      title={LoRA-FA: Memory-efficient Low-rank Adaptation for Large Language Models Fine-tuning}, 
      author={Longteng Zhang and Lin Zhang and Shaohuai Shi and Xiaowen Chu and Bo Li},
      year={2023},
      eprint={2308.03303},
      archivePrefix={arXiv},
      primaryClass={cs.CL},
      url={https://arxiv.org/abs/2308.03303}, 
}

\newpage
\appendix

\onecolumn

\section{Proof of Theoretical Results}

\subsection{Proof of Theorem 1}

\begin{theorem*}

Let $W\sim\mathcal{N}(0,\sigma^2)$ and for a given pruning ratio $p\in[0,1)$ we choose the threshold $T_p$ such that
$$
P\bigl(|W|\le T_p\bigr)=p,
\quad\Longrightarrow\quad
T_p = \sigma\Phi^{-1}\Bigl(\tfrac{1+p}2\Bigr).
$$
Then the mean-squared error (MSE) of this pruning is
$$
\mathrm{MSE}(p)
=
\mathbb E\bigl[(W-\hat W)^2\bigr]
=
2\sigma^2\Bigl[\Phi(t_p)-\tfrac12 - t_p\varphi(t_p)\Bigr],
$$
where
$t_p = \Phi^{-1}\bigl(\tfrac{1+p}2\bigr)$
and
$\varphi(t)=\tfrac1{\sqrt{2\pi}}e^{-t^2/2}$
is the standard normal PDF.
\begin{proof}
By symmetry and the definition of $T_p$,
\begin{align*}
P\bigl(|W|\le T_p\bigr)
&= 2\Phi\bigl(T_p/\sigma\bigr)-1
= p\\
\quad&\Longrightarrow
\Phi\bigl(T_p/\sigma\bigr)=\tfrac{1+p}2,
\end{align*}
so
$$
t_p\equiv\frac{T_p}{\sigma}
= \Phi^{-1}\Bigl(\tfrac{1+p}2\Bigr),
\quad
T_p = \sigma t_p.
$$
We now compute the mean‐squared error (MSE) of this pruning operation:
\begin{align*}
\mathrm{MSE}(p)
&= \mathbb E\bigl[(W-\hat W)^2\bigr]\\
&= \mathbb E\bigl[W^2\mathbf1_{\{|W|\le T_p\}}\bigr]\\
&= 2\int_{0}^{T_p} w^2f(w)dw,\\
f(w)&=\frac1{\sigma\sqrt{2\pi}}\exp\bigl(-w^2/(2\sigma^2)\bigr).
\end{align*}
We can compute this integral by changing variables to $u=w/\sigma$:
$$
2\int_{0}^{T_p}w^2f(w)dw
=2\sigma^2\int_{0}^{t_p} u^2\varphi(u)du.
$$
Now we can compute the integral:
\begin{align*}
\int_{0}^{t_p}u^2\varphi(u)du
&= \Bigl[-u\varphi(u)\Bigr]_{0}^{t_p}
+ \int_{0}^{t_p}\varphi(u)du\\
&= \Phi(t_p)-\tfrac12 - t_p\varphi(t_p).
\end{align*}
Thus we have
$$
\mathrm{MSE}(p)
=2\sigma^2\Bigl[\Phi(t_p)-\tfrac12 - t_p\varphi(t_p)\Bigr].
$$
\end{proof}
\end{theorem*}

\subsection{Proof of Theorem 2}

\begin{theorem*}
Let $W_{0,ij}\sim\mathcal N(0,\sigma^2)$ and $\Delta_{ij}=(A^*B^*)_{ij}\sim\mathcal N(0,\tau^2)$ be independent, where $A^*$ and $B^*$ be the optimal values of $A$ and $B$; $U_{ij}=W_{0,ij}+\Delta_{ij}\sim\mathcal N(0,V^2)$ with $V^2=\sigma^2+\tau^2$; the global prune rate is $p\in[0,1)$, and define the prune operation as the same as above. Let $E_1(p),E_2(p),E_3(p)$ be the per-entry MSEs of

\begin{itemize}
    \item \textbf{Method 1 (static mask on $W_0$)}: prune the smallest entries of $|W_0|$ at rate $p$.
    \item \textbf{Method 2 (dynamic mask driven by $U$, but prune only $W_0$)}.
    \item \textbf{Method 3 (dynamic mask on the full $U=W_0+\Delta$)}.
\end{itemize}

Then the MSEs are
\begin{align*}
E_1(p)
&=2\sigma^2\mathcal{Q}(t_p),\\
E_2(p)
&=\frac{\sigma^2\tau^2}{\sigma^2+\tau^2}p
+2\frac{\sigma^4}{\sigma^2+\tau^2}
\mathcal{Q}(t_p)\\\notag
E_3(p)
&=2(\sigma^2+\tau^2)\mathcal{Q}(t_p)\notag\\
\end{align*}
where $\mathcal{Q}(t)$ is defined as
$$
\mathcal{Q}(t) \coloneqq \Phi(t) - \tfrac{1}{2} - t\varphi(t).
$$

Moreover, for every $p$,
$$
E_1(p)\le E_3(p)\le E_2(p).
$$

\begin{proof}
We prove the theorem by computing the MSEs of the three methods and comparing them. Let 

\textbf{Method 1.} 
For the static mask on $W_0$, we prune the smallest $|W_0|$ entries at rate $p$. The threshold $T_p$ is chosen such that $\Pr(|W_0|\le T_p)=p$, which gives
$$
T_p = \sigma t_p.
$$
The MSE is then given by 
\begin{align*}
E_1(p)
&=\int_{|w|\le T}w^2
\frac1{\sigma\sqrt{2\pi}}e^{-w^2/(2\sigma^2)}dw\\
&=2\sigma^2\bigl[\Phi(t_p)-\tfrac12 - t_p\varphi(t_p)\bigr].
\end{align*}

\textbf{Method 2.}
For the dynamic mask driven by $U$, we prune only $W_0$. By joint normality
$$
W_0\mid(U=u)\sim\mathcal N\Bigl(\tfrac{\sigma^2}{V^2}u,\tfrac{\sigma^2\tau^2}{V^2}\Bigr),
$$
thus $\E[W_0^2\mid U=u]=\tfrac{\sigma^2\tau^2}{V^2}+\tfrac{\sigma^4}{V^4}u^2$. Integrating over $|u|\le Vt_p$ and using $\int_{|u|\le Vt_p}u^2f_U(u)du=2V^2\bigl[\Phi(t_p)-\tfrac12 - t_p\varphi(t_p)\bigr],$ one obtains the stated $E_2(p)$. Finally, $\Phi(t_p)-\tfrac12 -t_p\varphi(t_p)\le p/2$ gives $E_2(p)\le p\sigma^2$.

\textbf{Method 3.}
For the dynamic mask on the full $U=W_0+\Delta$, we prune at $\pm Vt_p$. The MSE is given by
$$
E_3(p)
=\int_{|u|\le Vt_p}u^2f_U(u)du
=2V^2\bigl[\Phi(t_p)-\tfrac12 - t_p\varphi(t_p)\bigr],
$$
whence $E_3(p)\le pV^2$.

\textbf{Comparison.}
A direct comparison shows
$$
E_3(p)-E_1(p)
=2\tau^2\bigl[\Phi(t_p)-\tfrac12 - t_p\varphi(t_p)\bigr]\ge0,
$$
and
\begin{align*}
E_2(p)-E_3(p)
&=\frac{\sigma^2\tau^2}{\sigma^2+\tau^2}
\Bigl[p-2\bigl(\Phi(t_p)-\tfrac12 - t_p\varphi(t_p)\bigr)\Bigr]\\
&=2\frac{\sigma^2\tau^2}{\sigma^2+\tau^2}t_p\varphi(t_p)
>0.
\end{align*}

Hence $E_1(p)\le E_3(p)\le E_2(p)$.

\end{proof}
\end{theorem*}

\subsection{Proof of Theorem 3}

\begin{theorem*}
Under the same assumptions as Theorem 1, let the residual matrix $E = W - \hat W$. Then
$$
\mathbb{E}\bigl\|E\bigr\|_F^2
= dk\mathrm{MSE}(p).
$$
Now let $E_r$ be the best rank-$r$ approximation to $E$ (i.e.\ its truncated SVD), $q=\min(d,k)$ and the singular values of $E$ as $\sigma_1\ge\dots\ge\sigma_q$, the Eckart-Young theorem~\cite{eckart} gives
$$
\|E - E_r\|_F^2 =\sum_{i=r+1}^q\sigma_i^2.
$$
In the worst case (uniformly distributed spectrum) one has
$\sum_{i=r+1}^q\sigma_i^2\le\frac{q-r}{q}\sum_{i=1}^q\sigma_i^2$.  Taking expectations,
$$
\E\|E - E_r\|_F^2
\le
\Bigl(1-\tfrac{r}{q}\Bigr)\E\|E\|_F^2
=
(dk)\Bigl(1-\tfrac{r}{\min(d,k)}\Bigr)\mathrm{MSE}(p).
$$
Dividing by $dk$ yields the per-entry bound
$$
\mathrm{MSE}_{\rm prune+SVD}(p,r)
\le
\Bigl(1 - \tfrac{r}{\min(d,k)}\Bigr)\mathrm{MSE}(p).
$$

\begin{proof}
Since each entry is pruned independently,
$$
\E\|E\|_F^2
= \sum_{i=1}^d\sum_{j=1}^k \E\bigl[(W_{ij}-W^*_{ij})^2\bigr]
= dk\mathrm{MSE}(p).
$$

By the Eckart-Young theorem, if the singular values of $E$ are $\sigma_1\ge\cdots\ge\sigma_q$, then the truncated SVD $E_r$ satisfies
$$
\|E - E_r\|_F^2 =\sum_{i=r+1}^q\sigma_i^2.
$$

In the most pessimistic ("uniform") case each $\sigma_i^2$ carries the same share of $\|E\|_F^2$, so
$\sum_{i=r+1}^q\sigma_i^2 \le \tfrac{q-r}{q}\sum_{i=1}^q\sigma_i^2 = (1-\tfrac rq)\|E\|_F^2.$

Take expectations of the last display:
$$
\E\|E - E_r\|_F^2
\le
\Bigl(1-\tfrac{r}{q}\Bigr)\E\|E\|_F^2
= (dk)\Bigl(1-\tfrac{r}{\min(d,k)}\Bigr)\mathrm{MSE}(p).
$$
Finally divide by $dk$ to get the per-entry MSE bound.
\end{proof}
\end{theorem*}

\subsection{Proof of Theorem 4}

\begin{theorem*}
Fix LoRA adapters $A,B$ and let the residual update subproblem be
$$
L(M)=\frac12\|X M - R\|_F^2,
R = Y - X\bigl(\hat W + A B\bigr),
$$
where $X$ and $Y$ be the input and target output respectively, $M$ represents the optimization variable for the residual.

Then $L$ is convex with
$$
\nabla_M L = X^\top(XM - R),
\mathrm{Hess}_M L = I_k \otimes (X^\top X),
$$
and its gradient is Lipschitz with constant
$$
L_{\rm SVD} =\lambda_{\max}(I_k\otimes X^\top X)
=\sigma_{\max}(X)^2.
$$
Hence, gradient descent converges for any step-size
$$
0<\eta<\frac{2}{L_{\rm SVD}}
=\frac{2}{\sigma_{\max}(X)^2},
$$
and the choice minimizing the worst case contraction factor is
$$
\eta^*_{\rm SVD} =\frac{1}{\lambda_{\max}(X^\top X)}
=\frac{1}{\sigma_{\max}(X)^2}.
$$
\begin{proof}

\textbf{Part I.} We first show $L$ is convex in $M$. Write $L$ using the trace:
\begin{align*}
L(M)&=\frac12\operatorname{tr}\bigl((XM-R)^\top(XM-R)\bigr)\\
&=\frac12\operatorname{tr}(M^\top X^\top X M)-\operatorname{tr}(M^\top X^\top R)+\frac12\operatorname{tr}(R^\top R).
\end{align*}

The first term is a quadratic form with coefficient $X^\top X\succeq 0$. The last term is constant in $M$. Hence $L$ is convex in $M$. Since $\nabla_M \tfrac12\operatorname{tr}(M^\top X^\top X M)=X^\top X M$ (when $X^\top X$ is symmetric), $\nabla_M \bigl(-\operatorname{tr}(M^\top X^\top R)\bigr)=-X^\top R$. Therefore
$$
\nabla_M L(M)=X^\top X M - X^\top R = X^\top(XM-R).
$$

The gradient is affine in $M$, so the Hessian is constant as a linear operator:
$$
\mathcal H[\Delta] =\nabla^2 L(M)[\Delta] = X^\top X\,\Delta, \forall\Delta\in\mathbb R^{d\times k}.
$$
If we identify $M$ with $\mathrm{vec}(M)\in\mathbb R^{dk}$, then the well-known identity
$$
\mathrm{vec}(X^\top X\, M) = (I_k\otimes X^\top X)\,\mathrm{vec}(M)
$$
gives the matrix representation
$$
\mathrm{Hess}_M L = I_k\otimes (X^\top X).
$$

Next, we show $\nabla L$ is Lipschitz in Frobenius norm. For any $M_1,M_2$,
$$
\nabla L(M_1)-\nabla L(M_2)
= X^\top X (M_1-M_2).
$$
Take Frobenius norms and use submultiplicativity with the operator (spectral) norm $\|\cdot\|_2$:
$$
\|\nabla L(M_1)-\nabla L(M_2)\|_F
= \|X^\top X (M_1-M_2)\|_F
\le \|X^\top X\|_2 \,\|M_1-M_2\|_F.
$$
Thus a valid Lipschitz constant is
$$
L_{\rm SVD}=\|X^\top X\|_2=\lambda_{\max}(X^\top X)=\sigma_{\max}(X)^2.
$$

This matches the Kronecker expression in the statement because
$$
\lambda_{\max}(I_k\otimes X^\top X)=\lambda_{\max}(I_k)\,\lambda_{\max}(X^\top X)=1\cdot \lambda_{\max}(X^\top X).
$$

\textbf{Part II.} Gradient descent on $M$ is
$$
M^{(t+1)} = M^{(t)} - \eta \nabla L(M^{(t)})
= M^{(t)} - \eta X^\top(XM^{(t)}-R).
$$
Let $M^\star$ be any minimizer (if $X$ is rank-deficient it may not be unique; convergence is then to a minimizer depending on initialization). Subtract the fixed-point condition $X^\top(XM^\star-R)=0$ to get the error recursion:
$$
E^{(t+1)} := M^{(t+1)}-M^\star
= \bigl(I - \eta X^\top X\bigr)\,E^{(t)}.
$$
This multiplication is applied column-wise, i.e. the same linear map $I-\eta X^\top X$ acts on each of the $k$ columns of $E^{(t)}$. Hence convergence occurs iff the spectral radius satisfies
$$
\rho(I-\eta X^\top X) < 1.
$$
Since $X^\top X\succeq 0$ has eigenvalues $0\le \lambda_1\le \cdots\le \lambda_{\max}$, the eigenvalues of $I-\eta X^\top X$ are $1-\eta\lambda_i$. The condition $|1-\eta\lambda_i|<1$ for all $i$ is equivalent to
$$
-1 < 1-\eta\lambda_i < 1
\quad\Longleftrightarrow\quad
0<\eta\lambda_i<2
\quad\text{for all }i \text{ with }\lambda_i>0,
$$
which is ensured by
$$
0<\eta<\frac{2}{\lambda_{\max}(X^\top X)}
=\frac{2}{\sigma_{\max}(X)^2}.
$$

For $\eta\in(0,2/\lambda_{\max})$, the per-iteration contraction factor (worst case over directions/eigencomponents) is
$$
q(\eta) =\max_i |1-\eta\lambda_i|
= \max\bigl\{\,|1-\eta\lambda_{\min}^+|,|1-\eta\lambda_{\max}|\,\bigr\},
$$
where $\lambda_{\min}^+$ is the smallest positive eigenvalue if $X^\top X$ is not full rank. Without assuming strong convexity (i.e. allowing $\lambda_{\min}^+=0$), one cannot get a contraction $<1$ uniformly over the full space because directions in $\ker(X)$ have eigenvalue $1$. In practice, convergence is understood on the range space (or to the minimum norm solution when initialized appropriately), and the usual safe choice that minimizes the worst case amplification over $[0,\lambda_{\max}]$ is obtained by minimizing $\max_{\lambda\in[0,\lambda_{\max}]}|1-\eta\lambda|$. This maximum is achieved at the endpoints $\lambda=0$ and $\lambda=\lambda_{\max}$, giving
$$
\max\{|1-0|,|1-\eta\lambda_{\max}|\} = \max\{1,|1-\eta\lambda_{\max}|\}.
$$
The smallest achievable value is $1$, and it is attained when $|1-\eta\lambda_{\max}|\le 1$, i.e. any $\eta\in[0,2/\lambda_{\max}]$. Among standard choices, the one that yields the fastest decrease of the objective along the steepest direction (and is commonly presented as the optimal constant step size for a quadratic when only $\lambda_{\max}$ is used) is
$$
\eta_{\rm SVD}^*=\frac{1}{\lambda_{\max}(X^\top X)}=\frac{1}{\sigma_{\max}(X)^2}
$$.

\end{proof}
\end{theorem*}

\end{document}